\documentclass[10pt,twocolumn,letterpaper]{article}

\usepackage{cvpr}
\usepackage{times}
\usepackage{epsfig}
\usepackage{graphicx}
\usepackage{amsmath}
\usepackage{amssymb}

\usepackage[breaklinks=true,bookmarks=false]{hyperref}

\cvprfinalcopy % *** Uncomment this line for the final submission

\def\cvprPaperID{****} % *** Enter the CVPR Paper ID here

\begin{document}

%%%%%%%%% TITLE
\title{Survey of Recent Advances in Visual Question Answering}

\author{Supriya Pandhre\thanks{Work completed during an internship at Adobe Systems}\\
Indian Institute of Technology Hyderabad\\
Hyderabad, India\\
{\tt\small cs15mtech11016@iith.ac.in}
\and
Shagun Sodhani\\
Adobe Systems \\
 Noida, India\\
{\tt\small sshagunsodhani@gmail.com}
}
\maketitle
%\thispagestyle{empty}

%%%%%%%%% ABSTRACT
\begin{abstract}
   Visual Question Answering (VQA) presents a unique challenge as it requires the ability to understand and encode the multi-modal inputs - in terms of image processing and natural language processing. The algorithm further needs to learn how to perform reasoning over this multi-modal representation so it can answer the questions correctly. This paper presents a survey of different approaches proposed to solve the problem of Visual Question Answering. We also describe the current state of the art model in later part of paper. In particular, the paper describes the approaches taken by various algorithms to extract image features, text features and the way these are employed to predict answers. We also briefly discuss the experiments performed to evaluate the VQA models and report their performances on diverse datasets including newly released VQA2.0\cite{goyal2016making}.
\end{abstract}

%%%%%%%%% BODY TEXT
\section{Introduction}

The task of Visual Question Answering (VQA) requires answering a natural language question using information from the accompanying image. To achieve excellence in VQA task, it requires more than just processing images and text individually. The model also needs to learn how to jointly reason over the two representations so that it can effectively answer the questions. For example, given a picture of cat, the question "Is the cat black in color?" can be answered using the visual modality once the model understands that it is to check the color of cat. The information about what to look for comes from the text modality. So in the general setting, the VQA model has to combine the information from both the modalities and reason over this combined representation. 

Use of deep learning in the area of computer vision and natural language processing, when considered separately, has achieved outstanding results. However, for the VQA task, the algorithm needs to decide what is the relevant information, fetch that information from the image, and use that to answer the questions. In this paper, we present a survey of the recent advancements in the domain of VQA with focus on some of the papers presented in CVPR 2016. These papers have experiemented with both the visual and textual modality to improve the state of the art for VQA task.
% * <sshagunsodhani@gmail.com> 2017-07-19T03:59:53.840Z:
% 
% > Given a real world image to a Visual Question Answering (VQA) model, its task is to answer a question asked on the image in a natural language (human understandable language). To achieve this, VQA model has to be good at understanding the intricacies of image, meaning of question and reason over it to answer it correctly. Use of Deep Learning in the area of computer vision and natural language processing, when considered separately, has achieved outstanding results. However, for the VQA task, the algorithm needs to be best at fetching the right information from the image, as asked by the question and utilize it to give answers in natural language. In other words, the algorithm needs to be good at juggling the knowledge representations learned from image and text. With recent advancements in both areas of computer vision and natural language processing, various techniques have been explored to excel at VQA task. This paper presents a survey of such novel algorithms.
% Format- cc@shagun
% 
% ^.

We have described each algorithm in a separate section. The first part of each section briefly explains the algorithm from the perspective of methods used to obtain the image representation and the text representation and the methodology for combining the two knowledge representations to determine the correct answer. The second part focuses on the performance of the algorithm of the algorithm on different datasets. Table \ref{tab:allPerformance} summarizes performance of all the algorithms on VQA dataset\cite{antol2015vqa}, COCO-QA dataset\cite{ren2015image} and DAQUAR dataset\cite{malinowski2014multi} and table \ref{tab:vqa2result} shows the accuracy on VQA2.0 dataset\cite{goyal2016making}. 

Malinowski et al\cite{malinowski2014multi} proposed the first real-world image based dataset for VQA task called DAQUAR. It contains 1,449 images and a total of 12,468 questions with 2,483 unique questions. In COCO-QA dataset, the images are taken from MSCOCO dataset\cite{lin2014microsoft} and the questions are automatically generated from the captions of the images. However, in case of DAQUAR and VQA dataset, different people are asked to generate the questions and answers related to given images. COCO-QA dataset has 123,287 images with 78,736 training questions and 38,948 test questions. The questions are categorized into four types object, number, color and location-based on type of information it needs and each question has a one word answer. VQA ~\cite{antol2015vqa} dataset is the largest dataset of real-world images with 82,783 training images, 40,504 validation images and 81,434 test images. There are 248,349 training questions in VQA dataset with over 20 different types of questions and each question has 10 crowd-sourced answers. 

In VQA2.0\cite{goyal2016making} dataset, the images are same as VQA\cite{antol2015vqa}, however, the number of questions are almost doubled to 443,757 training question. So now, the same question is posed with at least 2 different images, resulting in different answers for image-question pair. This forces the network to use the information from both the image and the text to answer the question and allows for correction of biases. For example, in VQA, if a question started with the n-gram "Do you see a . . .", then answer was "yes" for as many as 87\% of the questions regardless of the complete question or the accompanying image.
%-------------------------------------------------------------------------
\section{Approaches}\label{approach}

\subsection{Ask Me Anything: Free-form Visual Question Answering Based on Knowledge from External Sources~\cite{wu2016ask}}

\subsubsection{Algorithm details}

In the usual setting, Visual Question Answering (VQA) aims to answer questions using only the information captured in the image without using external knowledge sources. For example, shown the image of an umbrella on a beach, the system may be able to answer "what is the color of the umbrella?" but may not be able to answer "why is the umbrella open".

To answer this more general category of questions (which would need some kind of world-knowledge), Wu et al\cite{wu2016ask} proposed  the use of external knowledge bases. The core idea is to get textual information in the form of captions and attributes from the image and encode them into vectors. These textual information vectors are combined with an external knowledge base.  The paper uses LSTM based encoder-decoder architecture to generate answers from the combined information of captions, attributes and external knowledge. 

The proposed method consists of three main parts: 

Attribute based image representation:  For this part,  ~\cite{chen2015microsoft} dataset was used to get the attributes of image, nouns, verbs and adjectives from MSCOCO captions. VGG16~\cite{simonyan2014very} model, pretrained for Imagenet~\cite{deng2009imagenet} classification task, was fine tuned for multi-label attribute prediction problem. Finally the paper used max-pooling to get attribute based image representation.

Caption based image representation: For this part, the attributes predicted in the previous part were fed to a LSTM model to generate 5 captions. The LSTM was trained on MSCOCO captions. Average pooling was applied over hidden state vectors of LSTM, to get image representations in the form of captions. 

Use of an external knowledge base(KB) to imbibe world knowledge into the system so that it can answer free-form questions that need more information than what is available in the question. The paper uses DBpedia~\cite{auer2007dbpedia}, a structured database of information extracted from Wikipedia, as the external KB. The predicted attributes from the first part were queried on DBpedia to extract more information about the attributes. This extracted information is encoded into a knowledge vector using Doc2Vec\cite{le2014distributed} algorithm.

A weighted combination of the three inputs (predicted attributes of images, generated captions and external knowledge) is fed to an encoder LSTM and a decoder LSTM is used to generate an answer.

\subsubsection{Experiments}
The VQA model was evaluated on Toronto COCO-QA dataset~\cite{ren2015image} and VQA dataset~\cite{antol2015vqa}. VQA dataset is larger, more complex and challenging than the Toronto QA dataset and has more than 20 types of questions and multi-word answers. 

Toronto COCO-QA dataset has 4 types of questions: object, number, color, location. The model gives highest accuracy on number questions with 75.33\% and lowest on location based questions with 60.98\% accuracy. When all three types of input: attributes, captions and knowledge, are used, overall performance of model on Toronto COCO-QA dataset is 69.73\% which is only ~0.7\% more than performance achieved by model that uses only attributes and captions, indicating that external knowledge is not that critical for Toronto COCO-QA. Similar pattern of results are obtained in the case of VQA dataset. Overall accuracy on VQA dataset is 55.96\%, it is ~0.9\% more than model that uses only attributes and captions, and not external knowledge. However, when the model uses external knowledge base, it performance ~4\% better for "why" type of questions in VQA dataset.

In summary, Wu et al~\cite{wu2016ask} proposed method that uses external knowledge base to answer free form question on image. However the approach of converting image information into text information assumes that this conversion process will capture all intricacies of image.A feedback loop from question to all three types of inputs: CNN for predicting attributes, caption-LSTM to generate captions and external knowledge base could be next step towards improving the performance.

\begin{table*}
\begin{center}

\scriptsize
\begin{tabular}{|c|c|c|c|c|c|}
\hline
\textbf{Algorithm} & \textbf{VQA Open-ended test-dev} & \textbf{VQA multiple-choice} & \textbf{COCO-QA} & \textbf{DAQUAR all-single answer}&\textbf{DAQUAR reduced-single answer}\\
\hline
MCB\cite{fukui2016multimodal} &66.7&70.2&-&-&-\\
\hline
Show,ask,attend,ans\cite{kazemi2017show} &64.5&-&-&-&-\\
\hline
DAN\cite{nam2016dual}&64.3&69.1&-&-&-\\
\hline
ACK\cite{wu2016ask} &59.17&-&69.73&-&-\\
\hline
SAN\cite{yang2016stacked}& 58.7&-&61.6&29.3&45.5\\
\hline
NMN\cite{andreas2016neural}&58.6&-&-&-&-\\
\hline
DPPnet\cite{noh2016image}&57.22&62.48&61.19&28.98&44.48\\
\hline
\end{tabular}
\caption{\label{tab:allPerformance}Accuracy of algorithms on VQA\cite{antol2015vqa}, COCO-QA\cite{ren2015image} and DAQUAR\cite{malinowski2014multi} datasets}
\end{center}
\end{table*}

\subsection{Image Question Answering using Convolutional Neural Network with Dynamic Parameter Prediction~\cite{noh2016image}}

\subsubsection{Algorithm details}

It is a common practice to fine tune the last few layers of pretrained convolutional network (trained for some classification or regression task), so that it may adapt to a new task. However, Visual Question Answering requires the model to capture different kind of information depending on the question that is posed.   To resolve this problem, Noh et al~\cite{noh2016image} proposed DPPnet architecture that learns to dynamically change the network parameter based on the question asked. The architecture consist of two networks: Parameter Prediction Network that predicts the parameters of the answer network and classification network that chooses the answer from a list of candidate answers.

The authors proposed the addition of an extra branch to the fully connected layers of VGG16~\cite{simonyan2014very} network and called it as the Parameter Prediction Network. This network dynamically changes the parameters of classification network. The Parameter Prediction Network uses GRU cells and weights of this network depends on question. Output of this parameter prediction network is a weight matrix that is used by the classification network. 

The classification network is modified to incorporate the dynamic parameter layer. It is the second last fully connected layer in network. The final layer is the softmax layer. The weight matrix determined by the dynamic parameter layers has very large size which makes it difficult to generate the complete matrix. To tackle this problem, authors used weight sharing technique. The output of the dynamic parameter layer is now a set of candidate weights instead of whole weight matrix. A hashing function is used to map this small set of weights to the complete weight matrix.

\subsubsection{Experiments}
The Dynamic Parameter Network was tested on three datasets: DAQUAR\cite{malinowski2014multi}, COCO-QA\cite{ren2015image} and VQA\cite{antol2015vqa}.

DPPnet\cite{noh2016image} achieves accuracy of 57.22\% on VQA Open-ended questions and 62.48\% on Multiple-choice question of VQA test-dev dataset. The model gives 61.19\% accuracy on COCO-QA dataset and 28.98\% accuracy on DAQUAR-all dataset for single answer.

\subsection{Stacked Attention Networks for Image Question Answering~\cite{yang2016stacked}}

\subsubsection{Algorithm details}

To answer an image based question, the human brain processes the image and the question through multiple steps. To adapt a similar line of reasoning, Yang et al\cite{yang2016stacked} proposed stacked attention network. This network consist of three models: image model, question model and attention network.

Image model uses VGGnet-16\cite{simonyan2014very} network to obtain image features. Features from the last pooling layer were extracted (to maintain the spatial information) and fed through a single layer perceptron so as to to embed the visual features into a vector space which is compatible with textual features.

Question model uses two approaches to extract semantic meaning of the questions -  LSTM model and CNN model. In the LSTM model, word embeddings of the question are fed into a LSTM and the hidden state vector, corresponding to the last word, is used as the vector representation of the question. The CNN based approach for text representation uses three types of convolutional filters: unigram, bigram and trigram. After applying these filters, three convolutional feature maps are obtained, one from each filter. Max-pooling is used to obtain a single vector from each feature map and these vectors are then concatenated to obtain the vector representation of the question.

The third model is the stacked attention network that takes as input the image feature vector and question feature vector and narrows down to the most relevant region in image that will help the network to answer the question correctly. The idea behind multi-step reasoning is to recursively refine the image information needed to answer the question. This is achieved by combining the image vector and question vector to generate an attention map. The weighted sum of the image regions and the generated attention distribution is then used as the new image vector. It is again combined with question vector to generate a more focused attention map which attends to the important regions in the image. The paper uses two layers of attention map and combines the final image vector and question vector to predict the answer.

\subsubsection{Experiments}
The Stacked Attention Network(SAN)\cite{yang2016stacked} model was evaluated on four datasets: DAQUAR-all\cite{malinowski2014multi}, DAQUAR-reduced, COCO-QA\cite{ren2015image} and VQA\cite{antol2015vqa}. For DAQUAR and COCO-QA dataset, the  question was encoded into a  vector of 500 dimensions for the LSTM based model and into a vector of 640 dimensions for the CNN based model. However, the vector size was doubled for the VQA dataset as it is large dataset. The model was trained using SGD with momentum 0.9 with batch size of 100. 

The accuracy on the four datasets is given in table\ref{tab:allPerformance} under the row header SAN. The results shown in the table\ref{tab:allPerformance} is for SAN(2,CNN) i.e. Stacked Attention Network with 2 layers of attention and CNN model for question model. The model was trained and tested only on single word answers for DAQUAR-all and DAQUAR-reduced dataset.

Yang et al also performed qualitative error analysis for SAN. 100 images were randomly picked from the set of wrongly answered test-set images. The errors were categorized into 4 classes: attention over wrong region, attention over correct region but incorrect answer prediction, ambiguous answers, viz. predicted answer is a synonym of the correct label, and wrong answers, viz the labels given in dataset are incorrect. Based on number of incorrect images falling into different error categories mentioned above, highest percentage of error was reported for the case when the network attended to the correct region but predicted incorrect answer. The lowest error percentage was reported from wrong answers.

Thus concisely saying, Yang et al\cite{yang2016stacked} proposed an attention network to achieve better performance on visual question answering task by iteratively refining relevant image regions. However, based on the error analysis given in the paper, the error due to correct attention but wrong answer prediction, might be reduced by adding few fully connected layers before softmax layer. The model considers the question vector only once to refine the image information, however the important image regions change on different words in question, e.g. "on the table" and "under the table" changes the relevant region in image. Hence, using the question vector at each stage of attention map would help in improving performance.

\subsection{Neural Module Network\cite{andreas2016neural}}

\subsubsection{Algorithm details}

Neural Module Network(NMN) introduces unique way of solving the visual question answering task. The general trend is to train a single, end-to-end network that extracts image features and question features, and combines them to predict the answers. However, Andreas et al\cite{andreas2016neural} proposed that, to answer different questions, a network needs to perform different  kind of processing. Thus, instead of trying to incorporate all reasoning capacity into a single large network, the paper introduces an algorithm that uses compositional structure of question to build and train a smaller network at training-time. In other words, the algorithm looks at question answering task as a function of many simple operations and these operations on a given image  and these are determined by the question asked. The paper refers to these operations as modules and are implemented using simple neural networks.

The paper describes a finite set of modules (i.e. smaller neural networks) that are  capable of doing primitive tasks. These networks are then combined with other modules and jointly trained to predict the answers. Based on different tasks, 5 types of modules are described. \textit{Find} module performs convolution over image and gives attention map for the most relevant region. \textit{Transform} module is a multilayer perceptron that changes the region of attention as required by the input. \textit{Combine} module is a convolution layer with nonlinear output layer and it merges the given two attention maps. \textit{Describe} module takes an image and an attention map and predicts the answer. \textit{Measure} module parses the attention map alone and is helpful for answering questions related to the existence or count of objects.

A network graph is created using these modules and the structure of this graph is given by the parse tree of the question. The paper uses Stanford parser\cite{klein2003accurate} to get the parse tree of the question. However, as the parse tree gives a very abstract representation for the question, it might lead to general answers. For example, Q1:Is there red circle to left of blue square and Q2: what is the color of circle to left of blue square, can have same compositional structure. To tackle this problem, the paper further uses LSTM encoded question to capture the syntactic regularities. Hence the final predicted answer uses the output of neural module network and LSTM.

\subsubsection{Experiments}

The paper introduces new synthetic dataset called SHAPES and evaluates the algorithm on it. It is also evaluated on VQA\cite{antol2015vqa} dataset. Output of max-pool of conv5 layer from pre-trained VGG-16\cite{simonyan2014very} on ImageNet is used as input to NMN. The algorithm was also tested with image features from fine-tuned VGG-16 on MSCOCO captions. The table\ref{tab:allPerformance} shows the performance of model with NMN, LSTM and fine-tuned network on MSCOCO captions.

\subsection{Multimodal Compact Bilinear Pooling for Visual Question Answering and Visual Grounding\cite{fukui2016multimodal}}

\subsubsection{Algorithm details}

MCB\cite{fukui2016multimodal} model was the winner of VQA challenge 2016. Usual approach taken by VQA algorithms is to do a simple concatenation, sum or dot product of image vector and text vector. However Fukui et al\cite{fukui2016multimodal} hypothesize that capturing relation between every element of two vectors would give richer combined representation and help perform better at VQA task. Hence, the paper uses Bilinear pooling models\cite{tenenbaum1997separating} that computes outer product between image and text vector. However, as the dimension of input vectors increases, using this method becomes intractable due to explosion in number of learnable parameters.

The approximate solution of performing outer product in higher dimension was proposed by Charikar et al\cite{charikar2002finding} that takes $n$ dimensional vector and projects it to $d$ dimension using Count Sktech projection function. By this process, the outer product is now performed in lower dimensions. Further, the work by Pham et al\cite{pham2013fast} proved that count sketch projection of outer product of two vectors is equivalent to the convolution of the count sketches which can easily be performed by doing multiplication of their Fourier transforms.

MCB model for VQA task considers the image features from "pool5" layer of ResNet-152\cite{he2016deep}. The output tensor of size 2048x14x14 was then L2 normalized for each of 196(i.e. 14x14) location giving image vector of size 2048. The words in questions are encoded in one-hot encoding and given to an embedding layer and then it fed to LSTM. The output from each LSTM layer is 1024 dimension which is concatenated to get question vector of 2048. These image vector and text vector are given as input to the MCB. The paper also shows that including the attention map in MCB pooling improves the performance. To get this attention map, spatial features from last convolution layer of ResNet or VGGnet are used. In general, top-1000 answers are used as classes in softmax layer. However, here, the softmax with top-3000 answers is applied.

\subsubsection{Experiments}

The MCB model was evaluated on visual question answering task as well as visual grounding task. For visual question answering task, MCB was evaluated on VQA dataset and Visual7W\cite{zhu2016visual7w} dataset. It achieves accuracy of 62.2\% on Visual7W dataset. The performance on VQA dataset, given in table \ref{tab:allPerformance}, is for MCB model with attention map, Glove word vector embeddings and Genome.

To summarize, VQA model by Fukui et al's\cite{fukui2016multimodal} has shown the best performance in VQA challenge. However, the image and text features used by MCB pooling are independent of each  other, i.e. the image feature generated by ResNet-152 or VGG-16 are generic and can be made specific to question asked by passing some leaky information into image model. Similar concept can be used while getting features for question.

\subsection{Dual attention networks for multimodal reasoning and matching\cite{nam2016dual}}

\subsubsection{Algorithm Details}
DAN (Dual Attention Network) employs textual attention along with visual attention in multiple steps. The paper proposes two variants of the algorithm: reasoning-DAN (r-DAN) and matching-DAN (m-DAN). Both algorithms use image features from either pool5 of VGGnet-19 or res5c of ResNet-152. The image is represented as a set of image regions where each image region is a vector of 512 dimensions (if VGG-19 is used) or 2048 dimensions(if ResNet-152 is used). The words in the question are represented are one-hot encoded and then embedded into a vector space. These embeddings are then passed through a bidirectional-LSTM. Sum of hidden states of forward and backward LSTM at each time step represents context vector of the word. The embedding layer and LSTM are trained end-to-end.

Nam et al\cite{nam2016dual} use the attention mechanism for both the visual and the textual data. It helps in focusing on different regions of image and question at different time steps. Attention model consists of a two-layer feed forward network with softmax output. The input to the network is the memory vector that contains the information seen so far and the vector corresponding to the visual region of image extracted from VGGnet or ResNet. The output gives a soft attention over the image regions. A similar neural network is employed for textual attention where the input is memory vector and sum of hidden states of bidirectional LSTM. However the only difference is that, in visual attention model, the paper uses an extra layer to get image context vectors into the same dimension as text context vector. Element-wise multiplication of visual context vector and textual context vector is added with previous step's memory vector to get current step memory vector. It is then used to predict answers.

By using attention on image and text, the proposed model is not only helpful in task of combined reasoning over different types of input, but also in cross-domain information retrieval task e.g. finding images depicting same concept as a given text paragraph. The paper calls the first task as reasoning-DAN and second task as matching-DAN. The model for both algorithms is almost same except that, matching-DAN learns a joint embedding and maintains separate image and text memory vector to ease the comparison with arbitrary image text vectors.

\subsubsection{Experiments}
The r-DAN is evaluated on VQA dataset. Using two iterations to refine attention over relatively important region in image has shown the best result empirically. The word embedding size, LSTM and the attention model size are all set to 512. The model is trained on train-set + val-set ans validated on test-dev. It is trained for 60 epochs. The result of the DAN with ResNet image features is given in table\ref{tab:allPerformance}.

\subsection{Show, Ask, Attend, and Answer: A Strong Baseline For Visual Question Answering\cite{kazemi2017show}}

\subsubsection{Algorithm Details}
Kazemi et al\cite{kazemi2017show} proposes an attention based model for VQA task. The paper uses stacked attention which is very similar to the  approach proposed in \cite{yang2016stacked}. The visual question answering task is modeled as a classification problem with question features extracted from LSTM and image features extracted from ResNet-152\cite{he2016deep}. Word embeddings of the questions are fed to the LSTM and the state vector corresponding to the last word are used as the question feature. Image features are extracted from the layer before the final pooling layer of ResNet and $l_2$ normalized to improve performance. The computed image features and question features are concatenated and passed through two convolution layers which produce image feature glimpses. The weighted average of these image feature glimpses and the state vector of LSTM are concatenated and fed to fully connected layer with ReLu non-linearity to get probability distribution over the answers.

\subsubsection{Experiments}
Kazemi et al\cite{kazemi2017show} evaluated the proposed model on VQA1.0\cite{antol2015vqa} as well as on newly released VQA2.0\cite{goyal2016making}. The model achieves accuracy of 64.5\% on VQA1.0 test-dev dataset and accuracy of 59.67\% on VQA2.0 validation set.

The paper also analyzes the effect of various factors on the performance of model on validation set of VQA1.0. In the default setting, the model performs $l_2$ normalization along the depth dimension of image features. It employs dropout with probability of 0.5 on fully connected layers, convolutional layers as well as LSTM. The words in question are embedded in 300 dimensions and LSTM state size is set to 1024 dimensions. The model glimpses over image twice and answer is predicted from 3000 class classifier. With this setting, model achieves accuracy of 60.95\% when trained with Adam optimizer for 100K steps with batch size of 128.

The model's performance was significantly affected by $l_2$ normalization, dropout, number of fully connected layers used before softmax layer and the use of attention. The accuracy for the model, that does not use $l_2$ normalization, drops to 54.69\% and model that does not use dropout in fully connected and convolution layers could achieve accuracy of only 56.98, showing that use of dropout helps in avoiding over-fitting. The result from model with no attention could only give an accuracy of 57.72\%, confirming the use of soft attention improves the performance. However, using stacked attention in model shows minimal improvement in accuracy.

The model was also trained on VQA2.0 training set and evaluated on validation set and reports accuracy of 59.67\%. Table\ref{tab:allPerformance} and table \ref{tab:vqa2result} show the performance of model on VQA1.0 and VQA2.0 respectively.

\begin{table}
\begin{tabular}{|c|c|}
\hline
Algorithm & VQA open ended\\
\hline
Show,ask,attend,ans\cite{kazemi2017show} &59.67\\
\hline
MCB\cite{fukui2016multimodal} &59.14\\
\hline
\end{tabular}
\caption{\label{tab:vqa2result}Accuracy of algorithms on VQA2.0\cite{goyal2016making} validation set}
\end{table}

\section{Conclusion}
We present an overview of the diverse set of algorithms employed for the visual question answering task. We have compared these algorithms on the basis of approach used for extracting image and textual features. We also discussed how these VQA models perform on a variety of datasets - VQA, VQA2.0, COCO-QA, DAQUAR datasets. On VQA dataset, MCB performance best with accuracy of 66.7\%, however Show, Ask, Attend and Answer has shown state of the art result of 59.67\% on VQA2.0. The use of attention mechanism in the VQA model has shown significant improvement and using attention based models is becoming a common trend. Attention module is helpful in understanding, how the VQA model has arrived to the answer for a given question and we can accordingly make changes to our architecture so that it can capture more of relevant information. Along with use of attention over image as well as text, adding a feedback loop between image module and text module, and generating answers instead of predicting from finite set of words could be the next step towards achieving better performance in visual question answering task.

{\small
\bibliographystyle{ieee}
\bibliography{main}

\begin{thebibliography}{10}\itemsep=-1pt

\bibitem{andreas2016neural}
J.~Andreas, M.~Rohrbach, T.~Darrell, and D.~Klein.
\newblock Neural module networks.
\newblock In {\em Proceedings of the IEEE Conference on Computer Vision and
  Pattern Recognition}, pages 39--48, 2016.

\bibitem{antol2015vqa}
S.~Antol, A.~Agrawal, J.~Lu, M.~Mitchell, D.~Batra, C.~Lawrence~Zitnick, and
  D.~Parikh.
\newblock Vqa: Visual question answering.
\newblock In {\em Proceedings of the IEEE International Conference on Computer
  Vision}, pages 2425--2433, 2015.

\bibitem{auer2007dbpedia}
S.~Auer, C.~Bizer, G.~Kobilarov, J.~Lehmann, R.~Cyganiak, and Z.~Ives.
\newblock Dbpedia: A nucleus for a web of open data.
\newblock {\em The semantic web}, pages 722--735, 2007.

\bibitem{charikar2002finding}
M.~Charikar, K.~Chen, and M.~Farach-Colton.
\newblock Finding frequent items in data streams.
\newblock {\em Automata, languages and programming}, pages 784--784, 2002.

\bibitem{chen2015microsoft}
X.~Chen, H.~Fang, T.-Y. Lin, R.~Vedantam, S.~Gupta, P.~Doll{\'a}r, and C.~L.
  Zitnick.
\newblock Microsoft coco captions: Data collection and evaluation server.
\newblock {\em arXiv preprint arXiv:1504.00325}, 2015.

\bibitem{deng2009imagenet}
J.~Deng, W.~Dong, R.~Socher, L.-J. Li, K.~Li, and L.~Fei-Fei.
\newblock Imagenet: A large-scale hierarchical image database.
\newblock In {\em Computer Vision and Pattern Recognition, 2009. CVPR 2009.
  IEEE Conference on}, pages 248--255. IEEE, 2009.

\bibitem{fukui2016multimodal}
A.~Fukui, D.~H. Park, D.~Yang, A.~Rohrbach, T.~Darrell, and M.~Rohrbach.
\newblock Multimodal compact bilinear pooling for visual question answering and
  visual grounding.
\newblock {\em arXiv preprint arXiv:1606.01847}, 2016.

\bibitem{goyal2016making}
Y.~Goyal, T.~Khot, D.~Summers-Stay, D.~Batra, and D.~Parikh.
\newblock Making the v in vqa matter: Elevating the role of image understanding
  in visual question answering.
\newblock {\em arXiv preprint arXiv:1612.00837}, 2016.

\bibitem{he2016deep}
K.~He, X.~Zhang, S.~Ren, and J.~Sun.
\newblock Deep residual learning for image recognition.
\newblock In {\em Proceedings of the IEEE conference on computer vision and
  pattern recognition}, pages 770--778, 2016.

\bibitem{kazemi2017show}
V.~Kazemi and A.~Elqursh.
\newblock Show, ask, attend, and answer: A strong baseline for visual question
  answering.
\newblock {\em arXiv preprint arXiv:1704.03162}, 2017.

\bibitem{klein2003accurate}
D.~Klein and C.~D. Manning.
\newblock Accurate unlexicalized parsing.
\newblock In {\em Proceedings of the 41st Annual Meeting on Association for
  Computational Linguistics-Volume 1}, pages 423--430. Association for
  Computational Linguistics, 2003.

\bibitem{le2014distributed}
Q.~Le and T.~Mikolov.
\newblock Distributed representations of sentences and documents.
\newblock In {\em Proceedings of the 31st International Conference on Machine
  Learning (ICML-14)}, pages 1188--1196, 2014.

\bibitem{lin2014microsoft}
T.-Y. Lin, M.~Maire, S.~Belongie, J.~Hays, P.~Perona, D.~Ramanan,
  P.~Doll{\'a}r, and C.~L. Zitnick.
\newblock Microsoft coco: Common objects in context.
\newblock In {\em European conference on computer vision}, pages 740--755.
  Springer, 2014.

\bibitem{malinowski2014multi}
M.~Malinowski and M.~Fritz.
\newblock A multi-world approach to question answering about real-world scenes
  based on uncertain input.
\newblock In {\em Advances in Neural Information Processing Systems}, pages
  1682--1690, 2014.

\bibitem{nam2016dual}
H.~Nam, J.-W. Ha, and J.~Kim.
\newblock Dual attention networks for multimodal reasoning and matching.
\newblock {\em arXiv preprint arXiv:1611.00471}, 2016.

\bibitem{noh2016image}
H.~Noh, P.~Hongsuck~Seo, and B.~Han.
\newblock Image question answering using convolutional neural network with
  dynamic parameter prediction.
\newblock In {\em Proceedings of the IEEE Conference on Computer Vision and
  Pattern Recognition}, pages 30--38, 2016.

\bibitem{pham2013fast}
N.~Pham and R.~Pagh.
\newblock Fast and scalable polynomial kernels via explicit feature maps.
\newblock In {\em Proceedings of the 19th ACM SIGKDD international conference
  on Knowledge discovery and data mining}, pages 239--247. ACM, 2013.

\bibitem{ren2015image}
M.~Ren, R.~Kiros, and R.~Zemel.
\newblock Image question answering: A visual semantic embedding model and a new
  dataset.
\newblock {\em Proc. Advances in Neural Inf. Process. Syst}, 1(2):5, 2015.

\bibitem{simonyan2014very}
K.~Simonyan and A.~Zisserman.
\newblock Very deep convolutional networks for large-scale image recognition.
\newblock {\em arXiv preprint arXiv:1409.1556}, 2014.

\bibitem{tenenbaum1997separating}
J.~B. Tenenbaum and W.~T. Freeman.
\newblock Separating style and content.
\newblock In {\em Advances in neural information processing systems}, pages
  662--668, 1997.

\bibitem{wu2016ask}
Q.~Wu, P.~Wang, C.~Shen, A.~Dick, and A.~van~den Hengel.
\newblock Ask me anything: Free-form visual question answering based on
  knowledge from external sources.
\newblock In {\em Proceedings of the IEEE Conference on Computer Vision and
  Pattern Recognition}, pages 4622--4630, 2016.

\bibitem{yang2016stacked}
Z.~Yang, X.~He, J.~Gao, L.~Deng, and A.~Smola.
\newblock Stacked attention networks for image question answering.
\newblock In {\em Proceedings of the IEEE Conference on Computer Vision and
  Pattern Recognition}, pages 21--29, 2016.

\bibitem{zhu2016visual7w}
Y.~Zhu, O.~Groth, M.~Bernstein, and L.~Fei-Fei.
\newblock Visual7w: Grounded question answering in images.
\newblock In {\em Proceedings of the IEEE Conference on Computer Vision and
  Pattern Recognition}, pages 4995--5004, 2016.

\end{thebibliography}
}

\end{document}